# *Stitch, Contrast, and Segment*: Learning a Human Action Segmentation Model Using Trimmed Skeleton Videos

Haitao Tian[1*], Pierre Payeur[1]

[1]School of Electrical Engineering and Computer Science, University of Ottawa

htian026@uottawa.ca, ppayeur@uottawa.ca

## Abstract

Existing skeleton-based human action classification models rely on well-trimmed action-specific skeleton videos for both training and testing, precluding their scalability to real-world applications where untrimmed videos exhibiting concatenated actions are predominant. To overcome this limitation, recently introduced skeleton action segmentation models involve untrimmed skeleton videos into end-to-end training. The model is optimized to provide frame-wise predictions for any length of testing videos, simultaneously realizing action localization and classification. Yet, achieving such an improvement imposes frame-wise annotated skeleton videos, which remains time-consuming in practice. This paper features a novel framework for skeleton-based action segmentation trained on short trimmed skeleton videos, but that can run on longer untrimmed videos. The approach is implemented in three steps: *Stitch*, *Contrast*, and *Segment*. First, *Stitch* proposes a temporal skeleton stitching scheme that treats trimmed skeleton videos as elementary human motions that compose a semantic space and can be sampled to generate multi-action stitched sequences. *Contrast* learns contrastive representations from stitched sequences with a novel discrimination pretext task that enables a skeleton encoder to learn meaningful action-temporal contexts to improve action segmentation. Finally, *Segment* relates the proposed method to action segmentation by learning a segmentation layer while handling particular data availability. Experiments involve a trimmed source dataset and an untrimmed target dataset in an adaptation formulation for real-world skeleton-based human action segmentation to evaluate the effectiveness of the proposed method.

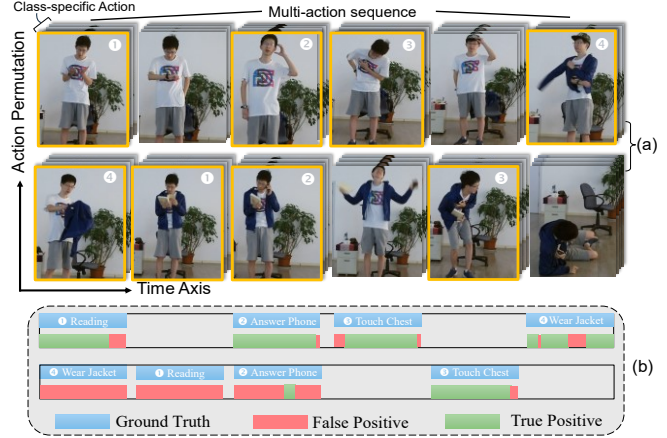

Figure 1: (a): Two video sequences from PKU MMD v2 exhibiting multiple actions that can be segmented into several class-specific actions. Four actions (highlighted with yellow outlines) found in both sequences are permutated in different orders of occurrence (1,2,3,4 vs. 4,1,2,3). (b): A well-trained[*] skeleton-based action segmentation model may generate very different segmentation results on the two sequences. We identify this issue as the shortcuts learning problem which results from learning action permutations during training. [*The MS-GCN (Filtjens et al. 2022) model was trained over the small PKU MMD v2 skeleton video-based dataset using the official Cross-View protocol after removing the training sequences with action permutation order: "4,1,2,3". More details are provided in the ablation studies section.]

## Introduction

Skeleton videos represent the trajectories of human body key joints captured by skeleton acquisition systems that leverage 2D and 3D computer vision with specialized software to estimate the spatial configuration of a human body skeleton model in a pre-calibrated environment. Such sensing systems offer the merit of being robust to variations in the environment compared to other modalities such as RGB frames (Carreira and Zisserman 2017) and optical flows (Feichtenhofer et al. 2016). Early research (Yan et al. 2018; Li et al. 2019; Chen et al. 2021) uses *trimmed* skeleton videos (i.e., class-specific action sequences as shown in Figure 1) to train human action recognition models, e.g., the end-to-end learning method (Yan et al. 2018) uses Graph Convolutional Networks (GCN) (Zhang et al. 2019) to parse human action dynamics from skeleton graphs, demonstrating high success rates in human-centered video analysis tasks (Niu et al. 2004). However, the scalability of existing human action recognition methods remains limited in the testing phase when an *untrimmed* skeleton video covering multiple actions in a sequence is captured by a skeleton acquisition system while the trained recognition model is tuned for single action prediction. To overcome the issue, recent research (Ma et al. 2021; Filtjens et al. 2022; Li et al. 2023; Yang et al. 2023; Chai et al. 2024) develop a skeleton-based action segmentation model, providing frame annotations for *untrimmed* videos as an input. The difficulty of this paradigm lies in the exploitation of temporal structures on untrimmed videos (as illustrated in Figure 1) that imposes the collection and annotation of a large untrimmed dataset, which is time-consuming and labor-intensive.

This work attempts to answer the following research question: **Is it possible to use trimmed skeleton videos to learn an effective action segmentation model that works on untrimmed videos?** In general, trimmed skeleton videos are easier to acquire and annotate from a controlled domain, e.g., public datasets or laboratory environments, offering a copious data volume for learning a real-world downstream task, e.g., action segmentation. An ideal scenario involves training an off-the-shelf segmentation model only using trimmed skeleton videos and making it applicable to a target application domain that involves untrimmed and unlabeled skeleton videos. Alternatively, one would pre-train an initial model upon trimmed videos and customize an action segmentation model for the target domain with a fine-tuning phase using a small number of untrimmed videos. Either strategy transfers skeleton action knowledge from a domain of low temporal granularity to that of high temporal granularity, effectively avoiding learning from scratch and reducing the annotation effort of videos in the target domain.

In this context, this work proposes: *Stitch, Contrast, and Segment*, as a temporal domain adaptation approach for skeleton-based action segmentation. (1) *Stitch*. The framework treats a dataset of trimmed skeleton sequences as a semantic space of elementary motions and uses frame correspondence to align short skeleton sequences on spatial trajectories. Skeleton sequences being robust to variations in the environment backgrounds, frame correspondence can be registered upon the spatial configurations of human skeletons (Figure 2 provides an illustration) to generate multi-action stitched skeleton sequences. (2) *Contrast*. The framework then considers the resulting stitched videos as an informative volume of rich temporal structures and exploits meaningful skeleton representations to improve action segmentation via pre-training. It differs from previous skeleton contrastive learning works that use trimmed sequences to design pretext tasks (Thoker et al. 2021; Guo et al. 2022; Huang et al. 2023; Zhang et al. 2023; Lin et al. 2023) or that use sliding windows to adapt sequence-wise classification to action segmentation (Chen et al. 2022; Lin et al. 2023; Yang et al. 2023). (3) *Segment*. Temporal action segmentation is challenging as it treats each frame in a sequence as an instance, which spirals the dimensionality of a network structure. To this end, we opt to use a modular action segmentation structure (Filtjens et al. 2022), which allows to adopt different learning strategies (e.g., supervised adaptation and zero-shot adaptation) towards real-world human action segmentation where the labeled data is scarce.

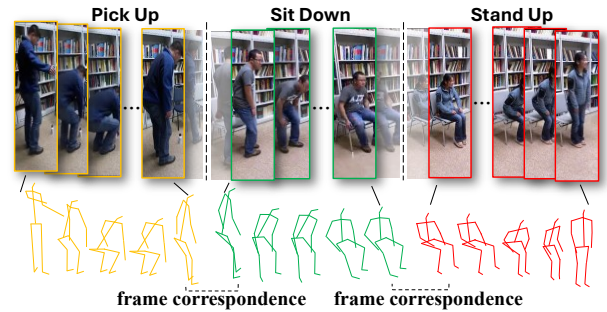

Figure 2: A concatenation of skeleton sequences resulting from stitching along the temporal axis. Upper row: RGB human motions involve strong dependency on environment backgrounds and significant variations in subject appearances. Lower row: skeleton motions represented by human key joints are agnostic to backgrounds, thus being easier to be manipulated with simple modifications (e.g., scaling and rotation). The proposed skeleton sequence stitching method involves using frame correspondence to align and concatenate short class-specific sequences.

The contributions of this work are threefold: *First*, it proposes a simple but effective temporal skeleton stitching method which can generate plausible untrimmed skeleton sequences where different actions are smoothly concatenated. *Second*, the proposed contrastive learning scheme demonstrates the effectiveness in exploiting temporal skeleton representations to improve action segmentation. *Lastly*, the novel temporal domain adaptation framework transfers skeleton action knowledge into real-world action segmentation which leads to significant reduction of the data collection and annotation requirements in the target domain. The project is available at: https://htian026.github.io/GCL, which contains supplementary material.

## Related Work

**Skeleton Action Recognition.** Previous works (Yan et al. 2018; Li et al. 2019; Chen et al. 2021) regard each skeleton sequence as a spatial-temporal graph and parse human action dynamics by spatial and temporal graph convolutions from massive skeleton data samples, thus realizing sequence-wise action classification for a given testing video. However, action recognition considers a simplified situation where both training and testing samples are trimmed to cover the entire movements involved in only one specific human action from the beginning to the end.

**Skeleton Action Segmentation.** Considering that skeleton videos from the real world are rarely segmented and typically display human actions in continuity, action segmentation is the natural evolution of action recognition. For instance, MS-GCN (Filtjens et al. 2022) is jointly learned over a GCN and a Temporal Convolutional Network (TCN)

(Carreira and Zisserman 2017)), where the former acts as a skeleton encoder to aggregate feature representations from untrimmed skeleton videos, while the latter models long-term action correlations from multi-scale temporal receptive fields. However, leveraging action segmentation escalates the difficulty of data acquisition as one needs to collect untrimmed long videos with frame-based annotation, which is labor intensive in practice.

**Transfer Learning in Action Segmentation.** Transfer learning algorithms help circumvent the shortage in data and have been used in RGB action video analysis. For example, I3D (Carreira and Zisserman 2017) is widely used in transfer learning for action segmentation by pre-training on short RGB videos (Piergiovanni and Ryoo 2018). In contrast, transfer learning in skeleton videos has not been extensively investigated. UNIK (Yang et al. 2021) involves pre-training a GCN on a large dataset and fine-tuning for action recognition in target domains with changing environments. To the best of our knowledge, the approach developed in the current paper presents the first method in transfer learning dedicated to skeleton action segmentation.

**Skeleton Data-based Contrastive Learning.** Contrastive learning has been utilized in skeleton representation learning (Thoker et al. 2021; Guo et al. 2022; Huang et al. 2023; Zhang et al. 2023; Lin et al. 2023). However, existing work focuses on action recognition and is difficult to apply to action segmentation. For instance, studies (Lin et al. 2023; Yang et al. 2023) utilize sliding windows to pre-segment an untrimmed sequence into equal-length sequence clips and apply representations extracted from trimmed skeleton videos to sequence clips. However, such a scheme essentially operates action recognition on untrimmed data, which inevitably introduces noisy boundaries on the segments and thus leads to compromised performance. In contrast, our work opts out of the use of sliding windows and formulates a contrastive learning framework specific for action segmentation.

## Problem Statement

### Skeleton-based Action Segmentation

A skeleton sequence, $L \in \mathbb{R}^{T \times V \times C_{in}}$, encodes the abstract dynamics of human movement over a period of time, $T$, depicting the trajectories of $V$ skeleton joints in $C_{in}$ (2 or 3) dimensions. When acquired with a skeleton acquisition system, a continuous (*untrimmed*) skeleton video tends to cover multiple actions concatenated in a single sequence.

**Definition 1** (*Action Permutation*) *Given an untrimmed skeleton video, $L \in \mathbb{R}^{T \times V \times C_{in}}$, we define the action permutation, $P = \{y_1, ..., y_n ..., y_N\}$, as a sequence indicating the order of occurrence of the actions within L, where $y_n$ is the action label of the n-th action's video segment $X_n \in \mathbb{R}^{T_n \times V \times C_{in}}$, with duration $T_n$ which leads to $T = \sum_{n=1}^{N} T_n$.*

*Herein, we have $L = X_1 \cdots \circ X_n \cdots \circ X_N$, where the symbol $\circ$ denotes the temporal concatenation operation.*

Conventional skeleton *action recognition* models favor single-action prediction (Kong and Fu 2022), i.e., $|P| = 1$. Aiming to process a multi-action input sequence, where $|P| > 1$, an *action segmentation* model involves frame-wise action categorization, which identifies the change of state between successive actions by assigning different action labels to specific frames in a skeleton video sequence.

### Temporal Domain Adaptation

Given that the annotation of untrimmed skeleton videos is extremely time-consuming and labor-intensive, this work focuses on a challenging alternative: learning skeleton-based action segmentation from trimmed skeleton videos. It assumes a large-scale source data domain, e.g., a laboratory dataset, containing well-trimmed and sequence-wise labeled short sequences $X$, and a small target data domain composed of untrimmed skeleton sequences $L$ (labeled or not). Our goal is to utilize the source data $X$ to help learn a robust and adaptive action segmentation model for the target domain associated with a few (if any) frame-wise labeled sequences within $L$. We call this process **temporal domain adaptation** as it transfers skeleton knowledge across different temporal granularities, e.g., from a coarse sequence-wise action recognition to finer frame-wise action segmentation. In this paper, two temporal domain adaptation objectives are pursued: supervised adaptation and zero-shot adaptation, which will be detailed in the next section.

## Methodology: Stitch, Contrast, and Segment

The proposed framework is structured in three steps. Initially, a skeleton temporal alignment and stitching strategy, *Skeleton-Stitch*, is proposed to generate stitched skeleton sequences. Second, it introduces Granular Contrastive Learning (*GCL*) to learn a skeleton encoder from stitched sequences. In the last step, it learns a segmentation layer upon the learnt skeleton encoder. The scalability of the proposed framework is demonstrated by applying *Skeleton-Stitch* and *GCL* to different learning strategies for achieving real-world skeleton-based action segmentation.

### Skeleton-Stitch on Trimmed Sequences

Observing that human actions are naturally and smoothly linked one to another when performed in a sequence, two skeleton frames on a temporal boundary within an untrimmed skeleton video, representing the end of the first action and the start of the second, shall be continuous in the location of their spatial joints. In this paper, we assume that if two skeleton frames, sampled from two distinct sequences, are aligned in spatial joint configurations, they could form some correspondence for stitching successive sequence segments into a continuous sequence.

Specifically, we present a simple but efficient strategy to realize temporal stitching of trimmed sequences. The goal is to generate a stitched sequence, $L$, conditioned by the action permutation, $P_L = \{y_1, ..., y_n, ..., y_N\}$, based on the given data batch $\mathbb{B}$, using a *Skeleton-Stitch* operation $\mathcal{T}$:

$$L = \mathcal{T}(\mathbb{B}, P) \quad (1)$$

where $\mathbb{B}$ is a disordered collection of trimmed skeleton videos. In Eq. (1), $\mathcal{T}$ accepts $\mathbb{B}$ as a semantic space for sampling and $P$ as a compositing surface for stitching. *Skeleton-Stitch* can be further decomposed as follows:

**Data Preprocessing.** We first apply a data pre-processing step to eliminate variation in views and scales. (1) We follow (Shi et al. 2019) to use a body coordinate system where the skeleton sequences are subtracted by the trajectory of the "spine" joint. (2) We follow (Liu et al. 2023) to apply a scaling operation that normalizes bone lengths on skeletons coming from different sequences.

**Definition 2** (*Frame Correspondence*) *Given two frames $p \in \mathbb{R}^{V \times C_{in}}$ and $q \in \mathbb{R}^{V \times C_{in}}$, sampled from two different skeleton sequences, $p$ is a frame correspondence of $q$, if their spatial distance $d(p, q) \leq d^*$. The metric $d$ is defined as $d(p, q) = \sum_{v=1}^{V} (||p_v - q_v||_2^2 + \max(1 - \frac{(p_v)^T q_v}{||p_v|| \, ||q_v||}, 0))$ which calculates the sum of the numerical differences on spatial locations and orientations at each skeleton joint. $d^*$ is a threshold to indicate a valid correspondence.*

**Frame Correspondence Registration on $\mathbb{B}$.** Given a template frame $p$ sampled from a sequence of $\mathbb{B}$, we register its frame correspondence as $q = \mathbb{B}[i, t]$, where

$$d(p, \mathbb{B}[i, t]) \leq d^* \quad (2)$$

where $t$ is the timestamp in the $i$-th sequence of $\mathbb{B}$. Eq. (2) encloses a loose constraint as $p$ may correspond to more than one frame correspondence given the variable value of $d^*$. To solve the ambiguity, we determine the correspondence by introducing additional computation constraints.

**Definition 3** (*Class-specific Subset*) *Let $B \equiv \{1, ..., |\mathbb{B}|\}$ be the indices of a batch of labeled data $\mathbb{B}$, we define the class-specific subset under the action category $k$ as $B_k \equiv \{i \in B: y_i = k\}$, where $y_i$ is the action label of $i$-th element in $\mathbb{B}$.*

**Multi-Action Stitches on $P$.** *Skeleton-Stitch* $\mathcal{T}$ uses $P = \{y_1, ..., y_n, ..., y_N\}$ as a compositing surface to stitch multi-action sequences that are located at class-specific subsets: $B_{y_1}, ..., B_{y_n}, ..., B_{y_N}$. Specifically, $\mathcal{T}$ implements a $N$-step recursive search. At step $n \geq 2$, it stitches the sequence

$$\tilde{L}_{(n)} = \tilde{L}_{(n-1)} \circ X_n \quad (3)$$

where $\tilde{L}_{(n-1)}$ is obtained at step $n-1$ with $P_{\tilde{L}_{(n-1)}} = \{y_1, ..., y_{n-1}\}$; $\tilde{L}_{(1)}$ is a single-action sequence randomly sampled from $B_{y_1}$. $X_n$ is a single-action segment with $P_{X_n} = y_n$, and is obtained by

$$X_n = \mathbb{B}[i, t:] \quad (4)$$
$$\text{s.t.} \quad i, t = \underset{\substack{i^* \in B_{y_n} \\ t^* \in \{1, ..., T^*\}}}{\arg\min} d(p, \mathbb{B}[i^*, t^*]) \quad (5)$$

Eq. (4) denotes that $X_n$ is a cropped sequence of $\mathbb{B}[i]$ by using the timestamps starting from $t$. Eq. (5) denotes that $i$ and $t$ are obtained by registering frame correspondence *w.r.t.* the template $p$ (defined as the last frame of $\tilde{L}_{(n-1)}$), where $i$ is searched from the class-specific subset $B_{y_n}$ and $t$ from the first $T^*$ timestamps of $X_i$ (where $T^* = |\mathbb{B}[i^*]|//\beta$, where $\beta$ acts as a threshold for controlling the search range). We solve Eq. (4) as a *0-1 Integer Programing* problem (details are available in supplementary material). The recursive operation stops at step $N$ and obtains a stitched sequence $L \equiv \tilde{L}_{(N)} = X_1 \cdots \circ X_n \cdots \circ X_N$.

## Granular Contrastive Learning on Stitched Sequences

In this section, we consider the stitched sequences generated by *Skeleton-Stitch* as an informative volume which can facilitate representation learning for action segmentation. Based on this, the framework proposes Granular Contrastive Learning (*GCL*), a novel contrastive learning framework compared to previous methods that use sliding windows.

**Skeleton Encoder.** We detail the proposed *GCL* framework in Fig. 3. It consists of a feature aggregation subnetwork, $\mathcal{M}: \mathbb{R}^{T \times V \times C_{in}} \to \mathbb{R}^{T \times C_{out}}$, parameterized by ST-GCN (Yan et al. 2018) which parses skeleton action encodings at the feature space, and a MLP-based feature projection layer, $\mathcal{G}: \mathbb{R}^{C_{out}} \to \mathbb{R}^{C_{prj}}$, providing one-dimensional vector of contrastive features at the output space.

**Data augmentation.** Previous contrastive learning paradigms utilize simple spatial and temporal skeleton modifications (Thoker et al. 2021; Guo et al. 2022; Huang et al. 2023; Zhang et al. 2023; Lin et al. 2023) as data augmentation tailored for independent single-action sequences, thus running short on exploiting contextual information between heterogenous actions. Aiming at representation learning for action segmentation, this work implements action-correlated data augmentations by means of *Skeleton-Stitch* $\mathcal{T}$.

**Definition 4** (*Granular Category Encodings*) *Given a stitched sequence, $L = X_1 \cdots \circ X_n \cdots \circ X_N$, and its activation at $\mathcal{M}$, i.e., $f = \mathcal{M}(L) \in \mathbb{R}^{T \times C_{out}}$, we treat $f$ as a concatenated sequence of categorical activations, i.e., $f = f_1 \cdots \circ f_n \cdots \circ f_N$, where $f_n \in \mathbb{R}^{T_n \times C_{out}}$ is the corresponding activation of $X_n$ whose action label is $y_n$ and duration is $T_n$. We have $T = \sum_{n=1}^{N} T_n$. We define the granular category encoding of $f_n$ as $h_n = \delta(f_n)$, where $\delta$ is a temporal pooling operation, i.e., $\delta: \mathbb{R}^{T \times C_{out}} \to \mathbb{R}^{C_{out}}$.*

Let $L_i = \mathcal{T}(\mathbb{B}, P_i)$ and $L_j = \mathcal{T}(\mathbb{B}, P_j)$ be two stitched sequences presenting different action permutations, i.e., $P_i \neq P_j$, and $k$ be one of the common action categories found in the two sequences, i.e., $k \in K \equiv (P_i \cap P_j)$, we use *granular*

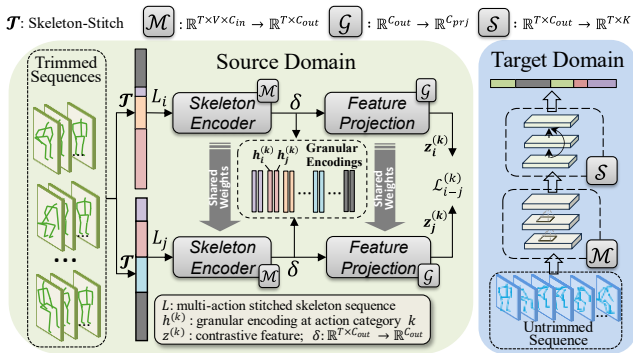

Figure 3: Conceptual overview of the proposed framework.

category encodings, $h_i^{(k)}$ and $h_j^{(k)}$, as a pair of *positives*, where $y_i = y_j = k$.

**Stitched sequence-based pretext task.** *GCL* aims to minimize the distance in between the positive pairs of all possible action category $k$ at each batch. Specifically, let $B_k$ be a class specific subset of granular category encodings of action category $k$ collected from multiple stitched sequences, it uses the projection layer, $\mathcal{G}$, to generate a pair of feature projections $z_i^{(k)} = \mathcal{G}(h_i^{(k)})$ and $z_j^{(k)} = \mathcal{G}(h_j^{(k)})$. The contrastive distance between $z_i^{(k)}$ and $z_j^{(k)}$ is calculated by the InfoNCE loss (Chen et al. 2020):

$$\mathcal{L}_{i-j}^{(k)} = -log \frac{\exp(z_i^{(k)} \cdot z_j^{(k)}/\tau)}{\exp(z_i^{(k)} \cdot z_j^{(k)}/\tau) + \sum_{m \in \mathbb{M}} \exp(z_i^{(k)} \cdot m/\tau)} \quad (6)$$

where $i \in B_k$, $j \in J_k \equiv B_k \setminus \{i\}$, and $k \in K$. In Eq. (6), $\tau$ acts as the temperature parameter, and $m$ denotes a *negative* counterpart sampled from a "memory bank", $\mathbb{M}$, which is used to compensate for using a small batch size due to memory limitation. We follow the same implementation (Guo et al. 2022) on the memory bank. The optimization upon Eq. (6) encourages the subnetwork $\mathcal{M}$ to represent closely aligned granular embeddings at the action category $k$. The final objective function is the sum of all positive samples at all action categories:

$$\mathcal{L} = \frac{1}{|K|} \sum_{k \in K} \frac{1}{|B_k|} \sum_{i \in B_k} \frac{1}{|J_k|} \sum_{j \in J_k} \mathcal{L}_{i-j}^{(k)} \quad (7)$$

We evaluate the skeleton representations of a pretrained subnetwork $\mathcal{M}$ in multiple adaptation tasks for action segmentation, which will be detailed in the next section.

## Skeleton Action Segmentation with Temporal Domain Adaptation

This section learns a segmentation (frame annotating) layer, $\mathcal{S}: \mathbb{R}^{T \times C_{out}} \rightarrow \mathbb{R}^{T \times K}$, which accepts hidden state at each frame from $\mathcal{M}$ as an instance and generates frame-wise softmax prediction for $K$ action categories (as illustrated in Figure 3). For implementation, we builds $\mathcal{S}$ upon a Temporal Convolutional Network (TCN) which discerns acausal action contexts (Lea et al. 2017) over adjacent frames by performing multi-scale dilated temporal convolutions (Farha and Gall 2019). We follow (Filtjens et al. 2022) for the implementation.

Our final action segmentation model is composed of the feature aggregation subnetwork, $\mathcal{M}$, and a segmentation layer, $\mathcal{S}$. Such a modularity allows us to relate the proposed *Skeleton-Stitch* and *Granular Contrastive Learning* to real-world action segmentation objectives while handling particular data availability. Particularly, it makes use of *source* trimmed sequences to pre-learn $\mathcal{M}$ via *GCL* and then realize multiple temporal domain adaptation strategies in the *target* domain. **In supervised adaptation**, it assumes that the *target* domain contains labeled untrimmed sequences, upon which we fine-tune $\mathcal{S}$ over the pre-trained $\mathcal{M}$. **In zero-shot adaptation**, as the *target* provides no training data, it uses *Skeleton-Stitch* to generate stitched data from the *source* domain for fine-tuning $\mathcal{S}$ over the pre-trained $\mathcal{M}$. More details on data generation are available in supplementary material.

# Experiments

In experiments, *Stitch, Contrast, and Segment* (SCS) involves pairs of datasets, in each case one trimmed and the other untrimmed, for each temporal domain adaptation task considered, i.e., supervised adaptation or zero-shot adaptation. Moreover, experimental ablation studies examine best practices in this work, facilitating its reproduction.

## Datasets

**NTU RGB+D** (Shahroudy et al. 2016) is a large-scale 3D human skeleton trimmed action dataset composed of 56,880 class-specific sequences. The dataset covers 60 human daily actions. The trimmed sequences are very short with low variation in action duration. **PKU-MMD** (Liu et al. 2017) is a popular 3D untrimmed skeleton dataset available in two versions. **PKU-MMD V1** covers 51 human daily actions. The dataset can be trimmed into 21,545 class-specific sequences. **PKU-MMD v2** contains 1,009 untrimmed video sequences covering 7,000 class-specific sequences. **Toyota Smarthome** (Das et al. 2019) is a large-scale 2D real-world video dataset for activities representing daily living. It covers up to 16,000 class-specific (trimmed) skeleton sequences with 31 action categories. **Toyota Smarthome Untrimmed** (Dai et al. 2022) is a real-world 2D dataset for daily living action segmentation. It contains 537 multi-action sequences covering 51 action categories.

## Implementation

In the experiments, we use multiple metrics to facilitate comparisons with related works. For instance, we use Top1 Accuracy (**Acc**) (Yan et al. 2018), mean Intersection over Union (**mIoU**) (Lin et al. 2023), mean Average Precision at IoU thresholds 0.1 (**mAP@0.1**) and 0.5 (**mAP@0.5**) (Yang et al. 2023), and per-frame mAP (**mAP$_f$**) (Dai et al. 2022) in action segmentation evaluation. Unless otherwise specified, all models are tested on the testing set of the target domain under the Cross-Subject protocol where the dataset is split upon subject identifications. We conduct **Linear** evaluation (where we fix $\mathcal{M}$ while fine-tuning $\mathcal{S}$) and **End-to-end** (**E2E**) evaluation (where we fine-tune $\mathcal{M}$ and $\mathcal{S}$ jointly) to ensure fair comparison with previous contrastive models. More details are available in supplementary material.

| NTU to PKU1 | Sliding Window | Evaluation Method | mAP% (0.1) | mAP% (0.5) |
|---|---|---|---|---|
| LAC (ICCV 23) | Y | Linear | 61.8 | - |
| **SCS (s.a.)** | N | Linear | **80.9** | **76.4** |
| CrosSCLR (CVPR 21) | Y | E2E | - | 60.1 |
| Hi-TRS (ECCV 22) | Y | E2E | - | 66.6 |
| LAC (ICCV 23) | Y | E2E | 92.6 | 90.6 |
| **SCS (s.a.)** | N | E2E | **94.6** | **91.1** |

Table 1: Experimental results with supervised temporal adaptation from "NTU RGB+D to PKU MMD v1".

| NTU to PKU2 | Sliding Window | Evaluation Method | mIoU% | Acc% |
|---|---|---|---|---|
| AimCLR (AAAI 22) | Y | Linear | 15.7 | 39.8 |
| ActCLR (CVPR 23) | Y | Linear | 21.4 | 51.3 |
| **SCS (s.a.)** | N | Linear | **53.4** | **69.5** |
| Hi-TRS (ECCV 22) | Y | E2E | - | 55.0 |
| **SCS (s.a.)** | N | E2E | **60.9** | **81.1** |

Table 2: Experimental results with supervised temporal adaptation from "NTU RGB+D to PKU MMD v2".

| Zero-Shot Adaptation | Evaluation Method | NTU to PKU1 mIoU% | NTU to PKU1 Acc% | NTU to PKU2 mIoU% | NTU to PKU2 Acc% |
|---|---|---|---|---|---|
| Baseline | E2E | 10.5 | 18.9 | 5.4 | 11.6 |
| **SCS (z.a.)** | Linear | 45.1 | 52.9 | **21.2** | 31.5 |
| **SCS (z.a.)** | E2E | **53.6** | **61.3** | 20.8 | **31.9** |

Table 3: Experimental results with zero-shot adaptation.

## Temporal Domain Adaptation Results

This section evaluates different learning strategies by amalgamating *Skeleton-Stitch* ($\mathcal{T}$) and *GCL* on multiple trimmed-untrimmed pairs of datasets. Our models are referred to as **SCS (s.a.)** in supervised adaptation and **SCS (z.a.)** in zero-shot adaptation.

**NTU Trimmed to PKU Untrimmed.** We consider a single source domain, NTU RGB+D, and two target domains, PKU-MMD v1 and PKU-MMD v2. **SCS (s.a.)** is compared to state-of-the-arts methods (Li et al. 2021; Thoker et al. 2021; Guo et al. 2022; Zhang et al. 2023; Lin et al. 2023; Yang et al. 2023) that utilize a sliding window to adapt contrastive skeleton representations from the trimmed source to the untrimmed target. Experimental results reported in Table 1 and Table 2 demonstrate that **SCS (s.a.)** significantly outperforms all state-of-the-art comparatives. We conclude that, by utilizing sliding window, existing work essentially operates action recognition (sequence-wise annotating) on the PKU-MMD domain from which the task granularity does not change across domains. In contrast, our model eliminates the use of a sliding window and learns a segmentation layer specific to the target domain, thereby leading to better performance. Second, **SCS (z.a.)** considers that the target domain is totally unknown, as we directly train an action segmentation model on the source domain NTU RGB+D. To the best of our knowledge, our method is the first to propose a trimmed-to-untrimmed zero-shot learning approach. For comparison, we learn baseline models using the original trimmed sequences from the source domain (all frames of each trimmed sequence are assigned the original sequence-wise action label). Table 3 presents the testing results, demonstrating the substantial benefit of learning from stitched sequences using the proposed method.

**PKU Trimmed to PKU Untrimmed.** Next, we consider the trimmed PKU-MMD v1 as the source domain and the untrimmed PKU-MMD v2 as the target domain. The skeleton sequences under the common 41 single-person actions are used for temporal domain adaptation training and testing. For comparison, we learn baseline models using the source trimmed sequences in zero-shot adaptation and using the target untrimmed sequences in supervised adaptation. Experimental results are summarized in the "**PKU1 to PKU2**" columns of Table 4 and demonstrate that the proposed models, **SCS (s.a.)** and **SCS (z.a.)**, outperform the baseline comparatives by large margins.

**TS Trimmed to TS Untrimmed.** The experiment **TST to TSU** considers the trimmed Toyota Smarthome (TST) as the source domain and the Toyota Smarthome Untrimmed (TSU) as the target domain. For the untrimmed sequences in the target domain, we map the frame annotations of each sequence that are not in the label space of the source domain as a background. Experimental results with respect to zero-shot adaptation and supervised adaptation are summarized in the last two columns of Table 4, also demonstrating significant improvement over baseline comparatives.

## Ablation Studies

Ablation experiments are conducted to better understand the effectiveness of each component introduced in this work.

**Stitched Data vs. Original Data**. *Skeleton-Stitch* uses frame correspondence to align heterogenous trimmed sequences and generate stitched sequences. In this section, we validate the effectiveness of the stitched sequences by reassembling datasets, e.g., PKU-MMD Untrimmed (v1 and v2) and Toyota Smarthome Untrimmed. Specifically, we first trim each sequence of the training set from an untrimmed dataset into class-specific sub-sequences using their associated data label and timestamps (we keep the original testing sets untouched for model evaluation). We shuffle the trimmed sequences and re-stitch into two restitched datasets, **Dataset I** (where we randomly group and concatenate trimmed sequences into restitched sequences without registering frame correspondence) and the **Dataset II** (where we use $\mathcal{T}$ to stitch sequences upon frame correspondence). The restitched datasets have the same size (i.e., the number of data samples) and same semantic information (i.e., the action permutations of each sequence) as the original dataset counterpart. The performance of trained action segmentation models is demonstrated in Figure 4 (a). First, training over a randomly stitched dataset, **Dataset I**, leads to drastic performance decrease compared to the baseline model that uses the original datasets. In contrast, **Dataset II** achieves comparable performance to the baseline over all three untrimmed datasets (PKU1, PKU2, TSU), suggesting that the *Skeleton-Stitch* method is able to generate plausible sequences in line with the contents of original data.

| Models | Eval. Method | Sou. Data | Tar. Data | GCL | $\mathcal{T}$ | PKU1 to PKU2 mIoU% | Acc% | TST to TSU mAP$_f$% | Acc% |
|---|---|---|---|---|---|---|---|---|---|
| Baseline | E2E | - | ✓ | - | - | 51.8 | 68.1 | 24.1 | 20.6 |
| SCS (s.a.) | Linear | ✓ | ✓ | ✓ | ✓ | 64.0 | 74.9 | 28.3 | 26.7 |
| SCS (s.a.) | E2E | ✓ | ✓ | ✓ | ✓ | 71.4 | 84.6 | 35.1 | 33.2 |
| Baseline | E2E | ✓ | - | - | - | 10.1 | 18.6 | 1.6 | 4.8 |
| SCS (z.a.) | Linear | ✓ | - | ✓ | ✓ | 24.9 | 36.7 | 20.5 | 19.4 |
| SCS (z.a.) | E2E | ✓ | - | ✓ | ✓ | 25.5 | 38.2 | 21.2 | 20.9 |

Table 4. Action segmentation results with temporal domain adaptation. Models are tested on the respective original testing set of the target domain under the Cross-Subject protocol.

**Data Augmentation**. The use of stitched data is also investigated as a source for data augmentation in both end-to-end supervised learning and pretext task-based contrastive learning. More specifically, it enables to augment a small untrimmed dataset in an efficient way. **Expanded Dataset** in Figure 4 (a) demonstrates that after combining the original dataset with **Dataset II**, substantial improvement is obtained over the three untrimmed datasets. Second, we verify the effectiveness of stitched data in contrastive learning, i.e., the *GCL* model in Figure 4 (b). For comparison, we also use classical data augmentation strategies: spatial transformation and temporal scaling (Guo et al. 2022), to develop a baseline model. The experimental comparison between the baseline and *GCL* in Figure 4 (b) suggests that classical augmentation favoured by previous works may lead to a limited transferability in the adaptation towards action segmentation. However, normal classical augmentations are orthogonal to our method. After amalgamating the two types of data augmentations (*Classical Aug.*+$\mathcal{T}$), the adaptation performance on all datasets is effectively boosted.

**Cross-Sequence Test.** In this section, we extend the discussion initiated in Figure 1. PKU-MMD v2 is a small dataset covering 26 action permutations that are commonly shared by training and testing samples under the Cross-View protocol (where data is split based on camera views). We follow a custom Cross-Sequence protocol for the dataset where the training data covers 15 action permutations, and the testing data involves the remaining 11 permutations. We train models with two datasets: the original dataset and the expanded dataset using stitched sequences generated by *Skeleton-Stitch* (detailed in Supplementary Material). For comparison, we also provide results *w.r.t.* the Cross-View protocol. Figure 4 (c) presents the experimental results. It shows generalization discrepancy between action segmentation models evaluated under different evaluation protocols. **Upper row**: the baseline modes that are trained with original datasets demonstrate poor generalization (gap between training and testing accuracies) capability under the Cross-Sequence protocol. We conjuncture that as the action permutation information is shared between the training data and the testing data under the Cross-View protocol, the baseline model can use observed action permutation as a shortcut to solve the segmentations on the testing data. Therefore, the model fails in the Cross-Sequence protocol which occludes the sharing of action permutation. **Lower row**: the enhanced models are trained with the expanded datasets and achieve relatively stable generalization performance in both the cross-view and cross-sequence protocols after being exposed to more action contexts via the use of *Skeleton-Stitch*.

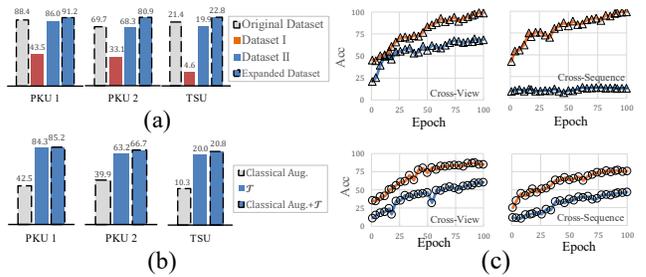

Figure 4: Effectiveness (Acc%) of Skeleton-Stitch in (a) end-to-end supervised learning, (b) GCL based pre-training; (c) Generalization performance without (△) and with (◦) stitched sequences. Orange lines denote training accuracy across epochs and blue lines relate to test accuracy.

## Conclusion

This paper builds upon the observation that the effectiveness of existing skeleton-based action recognition models is significantly uprooted when handling naturally continuous transitions between human actions in an untrimmed skeleton video, while learning segmentation models imposes expensive requirements for frame-wise data annotation. This paper introduces a temporal domain adaptation framework for action segmentation by treating trimmed skeleton videos as a research instance. It first proposes to generate spatially continuous and stitched skeleton sequences where different actions are smoothly concatenated. Second, it utilizes contrastive learning to learn fine-grained temporal segmentation embeddings from a source domain that can be represented by a skeleton encoder to transfer knowledge into a target domain for the task of action segmentation. The scalability of the proposed method is validated by comprehensive experimental results in supervised and zero-shot adaptation.

# Acknowledgements

This research was supported in part by MITACS Accelerate and NSERC Discovery grants. The authors also acknowledge the collaboration of Spectronix Inc.
# References

Cao, Z.; Simon, T.; Wei, S.-E.; and Sheikh, Y. 2017. Realtime multi-person 2d pose estimation using part affinity fields. *Proceedings IEEE Conference on Computer Vision and Pattern Recognition*, 7291–7299.

Carreira, J.; and Zisserman, A. 2017. Quo vadis, action recognition? a new model and the kinetics dataset. *Proceedings IEEE Conference on Computer Vision Pattern Recognition*, 6299–6308.

Chai, S.; Jain, R. K.; Liu, J.; Teng, S.; Tateyama, T.; Li, Y.; and Chen, Y.-W. 2024. A motion-aware and temporal-enhanced Spatial–Temporal Graph Convolutional Network for skeleton-based human action segmentation. *Neurocomputing*, 580, 127482.

Chen, X.; Fan, H.; Girshick, R.; and He, K. 2020. Improved baselines with momentum contrastive learning. *arXivpreprintarXiv:2003.04297*.

Chen, Y.; Zhang, Z.; Yuan, C.; Li, B.; Deng, Y.; and Hu, W. 2021. Channel-wise topology refinement graph convolution for skeleton-based action recognition. *Proceedings IEEE/CVF International Conference on Computer Vision*, 13359–13368.

Chen, Y.; Zhao, L.; Yuan, J.; Tian, Y.; Xia, Z.; Geng, S.; Han, L.; and Metaxas, D. N. 2022. Hierarchically self-supervised transformer for human skeleton representation learning. *European Conference Computer Vision*, 185–202.

Dai, R.; Das, S.; Sharma, S.; Minciullo, L.; Garattoni, L.; Bremond, F.; and Francesca, G. 2022. Toyota smarthome untrimmed: Real-world untrimmed videos for activity detection. *IEEE Transactions Pattern Analysis Machine Intelligence*, 45(2), 2.

Das, S.; Dai, R.; Koperski, M.; Minciullo, L.; Garattoni, L.; Bremond, F.; and Francesca, G. 2019. Toyota smarthome: Real-world activities of daily living. *Proceedings IEEE/CVF International Conference on Computer Vision*, 833–842.

Farha, Y. A.; and Gall, J. 2019. Ms-tcn: Multi-stage temporal convolutional network for action segmentation. *Proceedings IEEE/CVF Conference on Computer Vision and Pattern Recognition*, 3575–3584.

Feichtenhofer, C.; Pinz, A.; and Zisserman, A. 2016. Convolutional two-stream network fusion for video action recognition. *Proceedings IEEE Conference on Computer Vision and Pattern Recognition*, 1933–1941.

Filtjens, B.; Vanrumste, B.; and Slaets, P. 2022. Skeleton-based action segmentation with multi-stage spatial-temporal graph convolutional neural networks. *IEEE Transactions Emerging Topics Computing*, 12(1), 1.

Guo, T.; Liu, H.; Chen, Z.; Liu, M.; Wang, T.; and Ding, R. 2022. Contrastive learning from extremely augmented skeleton sequences for self-supervised action recognition. *Proceedings AAAI Conference on Artificial Intelligence*, 36(1), 1.

Huang, X.; Zhou, H.; Wang, J.; Feng, H.; Han, J.; Ding, E.; Wang, J.; Wang, X.; Liu, W.; and Feng, B. 2023. Graph contrastive learning for skeleton-based action recognition. *arXiv preprint arXiv:2301.10900*.

Kong, Y.; and Fu, Y. 2022. Human action recognition and prediction: A survey. *International Journal Computer Vision*, 130(5), 5.

Lea, C.; Flynn, M. D.; Vidal, R.; Reiter, A.; and Hager, G. D. 2017. Temporal convolutional networks for action segmentation and detection. *Proceedings IEEE Conference on Computer Vision Pattern Recognition*, 156–165.

Li, M.; Chen, S.; Chen, X.; Zhang, Y.; Wang, Y.; and Tian, Q. 2019. Actional-structural graph convolutional networks for skeleton-based action recognition. *Proceedings IEEE/CVF Conference on Computer Vision and Pattern Recognition*, 3595–3603.

Li, Y.; Li, Z.; Gao, S.; Wang, Q.; Hou, Q.; and Cheng, M.-M. 2023. A Decoupled Spatio-Temporal Framework for Skeleton-based Action Segmentation. *arXiv preprint arXiv:2312.05830*.

Li, L.; Wang, M.; Ni, B.; Wang, H.; Yang, J.; and Zhang, W. 2021. 3d human action representation learning via cross-view consistency pursuit. *Proceedings IEEE/CVF Conference on Computer Vision and Pattern Recognition*, 4741–4750.

Lin, L.; Zhang, J.; and Liu, J. 2023. Actionlet-dependent contrastive learning for unsupervised skeleton-based action recognition. *Proceedings IEEE/CVF Conference on Computer Vision and Pattern Recognition*, 2363–2372.

Liu, C.; Hu, Y.; Li, Y.; Song, S.; and Liu, J. 2017. Pku-mmd: A large scale benchmark for continuous multi-modal human action understanding. *arXivpreprintarXiv:1703.07475*.

Liu, H.; Liu, Y.; Mu, T.-J.; Huang, X.; and Hu, S.-M. 2023. Skeleton-CutMix: Mixing Up Skeleton with Probabilistic Bone Exchange for Supervised Domain Adaptation. *IEEE Transactions Image Processing*.

Ma, H.; Yang, Z.; and Liu, H. 2021. Fine-grained unsupervised temporal action segmentation and distributed representation for skeleton-based human motion analysis. *IEEE Transactions Cybernetics*, 52(12), 12.

Niu, W.; Long, J.; Han, D.; and Wang, Y.-F. 2004. Human activity detection and recognition for video surveillance. *2004 IEEE International Conference Multimedia Expo (ICME) (IEEE Cat. No. 04TH8763)*, 1, 719–722.